\documentclass[letterpaper, preprint, paper,11pt]{AAS}	

\usepackage{bm}
\usepackage{amsmath}
\usepackage{subfigure}
\usepackage[colorlinks=true, pdfstartview=FitV, linkcolor=black, citecolor= black, urlcolor= black]{hyperref}
\usepackage{overcite}
\usepackage{footnpag}			      	
\usepackage{todonotes}

\PaperNumber{22-811}

\begin{document}

\title{Detection and Initial Assessment of Lunar Landing Sites Using Neural Networks}

\author{Daniel Posada\thanks{Ph.D. Candidate, Space Technologies Lab, Embry-Riddle Aeronautical University, 1 Aerospace Blvd., Daytona Beach, FL, 32114.},  
Jarred Jordan\thanks{Undergraduate, Space Technologies Lab, Embry-Riddle Aeronautical University, 1 Aerospace Blvd., Daytona Beach, FL, 32114.}, Angelica Radulovic\footnotemark[2],
Lillian Hong\thanks{Graphics Programmer, Intuitive Machines, 3700 Bay Area Blvd, Houston, TX, 77058.},
Aryslan Malik\thanks{Visiting Assistant Professor, Space Technologies Lab, Embry-Riddle Aeronautical University, 1 Aerospace Blvd., Daytona Beach, FL, 32114.},
\ and Troy Henderson\thanks{Associate Professor, Space Technologies Lab, Embry-Riddle Aeronautical University, 1 Aerospace Blvd., Daytona Beach, FL, 32114.}
}

\maketitle{}

\begin{abstract}
Robotic and human lunar landings are a focus of future NASA missions. 
Precision landing capabilities are vital to guarantee the success of the mission, and the safety of the lander and crew.
During the approach to the surface there are multiple challenges associated with Hazard Relative Navigation to ensure safe landings.
This paper will focus on a passive autonomous hazard detection and avoidance sub-system to generate an initial assessment of possible landing regions for the guidance system.
The system uses a single camera and the MobileNetV2 neural network architecture to detect and discern between safe landing sites and hazards such as rocks, shadows, and craters. Then a monocular structure from motion will recreate the surface to provide slope and roughness analysis.
\end{abstract}

\section{Introduction}
With NASA's recent emphasis on lunar missions, both private companies and space agencies are planning missions to go back to the Moon. There is a growing need for space hardware and software capable of guaranteeing mission success. The first missions to return to the Moon since Apollo are part of the Commercial Lunar Payload Services (CLPS) awards by the National Aeronautics and Space Administration (NASA)\cite{Daines_2019}. Then, following the Artemis program, astronauts will land with the recently awarded Human Landing System (HLS) spacecraft. There are an expected 90 missions to the Moon by the year  2030.\cite{Balossino_2022} These missions undergo many challenges for landing; to illustrate the difficulty, some recent attempts that failed are the Indian Space Research Organisation Vikram and the Israel Aerospace Industries Beresheet lunar lander. Landing a spacecraft on an unknown surface with unknown hazards is risky; therefore, the terminal phase of any space exploration mission is a critical moment.
This stage will likely be the closest the lander will be to experience any physical hazard as it is performing a touchdown maneuver.

There are multiple active and passive sensors that can provide awareness for this purpose. For example Light Detection and Ranging (LiDARs) which are active sensors provide high fidelity mapping however they are still being designed for space, are expensive, power-hungry, and bulky. Cameras on the contrary, are great passive sensors due to the amount of information they provide, and their footprint on a vehicle is smaller. However, the downfall of these two sensors is the amount of data to be processed, and interpretation is as good as the processing algorithms. Additionally, cameras require light (either natural or provided) in order to operate, potentially limiting future landing site areas.

An image is composed of millions of pixels that translates into processing very large arrays, and performing any computer vision task implies operating over this amount of data. To provide awareness of the desired landing sites, a deep learning architecture was selected due to its benefits in processing data sets like images in a fast fashion by reducing dimensions. MobileNetV2 is a well-known deep learning architecture implemented for multiple classification, detection, and segmentation tasks \cite{mobilenetv2}. This network was designed and optimized to be used in a mobile environment providing a useful portable method without the need of powerful computation. This pipeline provides an easy solution to be implement in robotic spacecraft lander due to cost, optimization, and size. This paper will assume a small lunar lander with a Vehicle Footprint Dispersion Ellipse (VFDE) of 10 meters. In other words, the size of the spacecraft and the GNC error is accounted for within this footprint. The VFDE also provides valuable information as it will provide the minimum area to be analyzed that it is considered safe for the landing.

\section{Hazard Detection and Avoidance}
There has been some previous Hazard Detection and Avoidance (HDA) development efforts\cite{epp2007autonomous,crane2014vision,yu2014new,jiang2016innovative, iiyama2021deep, tomita2022sequential}, but only a few have really been implemented taking into account that any HDA implementation is unique to the landing requirements\cite{posada2020a,getch2022}, available sensors, and computational requirements\cite{villalpando2010investigation,johnson2010analysis} of the mission. The only current active implementation of an HDA development was used for the landing of Mars 2020 Perseverance Rover \cite{aaron2022camera}. This landing combined their Terrain Relative Navigation (TRN) technology with a camera that resized the imagery to process it in time against a pre-loaded map that required time for development using the Mars Reconnaissance Orbiter\cite{cheng2021making}. These systems require more testing but they will become more common with the increase in missions. Other newer approaches take use the footprint of the hazards of landing site and the geometry of the spacecraft, but require significant computational capabilities\cite{nelson2021landing}. 
Previous research and current state-of-the-art mostly assume that LiDAR technology is ready for spaceflight\cite{johnson2002lidar,moghe2020deep,iiyama2021deep}; however, this active sensor is still being developed for this application (power consumption, mass, and volume) and has a long way to be tested\cite{amzajerdian2013lidar,roback2013helicopter}. Therefore, passive sensors such as cameras are still one the most viable sensors for this application with the known disadvantage of the sensitivity to the light conditions.

One of the biggest challenges of performing Hazard Relative Navigation (HRN) is to be able to perform HDA in almost real-time. Being able to provide the Guidance, Navigation, and Control (GNC) system with a waypoint for landing requires multiple steps. Classical computer vision methods require significant computational power in order to process the amount of data and the operations performed in these arrays require time, for this reason images are down-sampled (resized) to improve speed at the expense of accuracy. An example is detection and segmentation of rocks, shadows, and craters.
These task requires three independent operations tailored to the properties of each hazard which requires intensive computations. However, the use of MobileNetV2 brings a direct operation on the image that can detect all of these hazards at the same time.
This is done in an efficient way with great accuracy by mapping the features detected from the image into different higher dimensions to extract their unique features making it more robust to different lighting conditions, as it employs less hyper-parameters and can be run on computationally-constrained mobile hardware providing almost real-time inference which makes it an effective auxiliary tool for safe landing. Other approaches to this problem using semantic segmentation and reinforced learning have been proposed recently. These papers illustrate the challenges of a mobile environments and provides an alternative on the use of NNs to solve the problem\cite{tomita2020hazard,tomita2022adaptive}.

\begin{figure}[htb]
	\centering\includegraphics[width=0.85\textwidth]{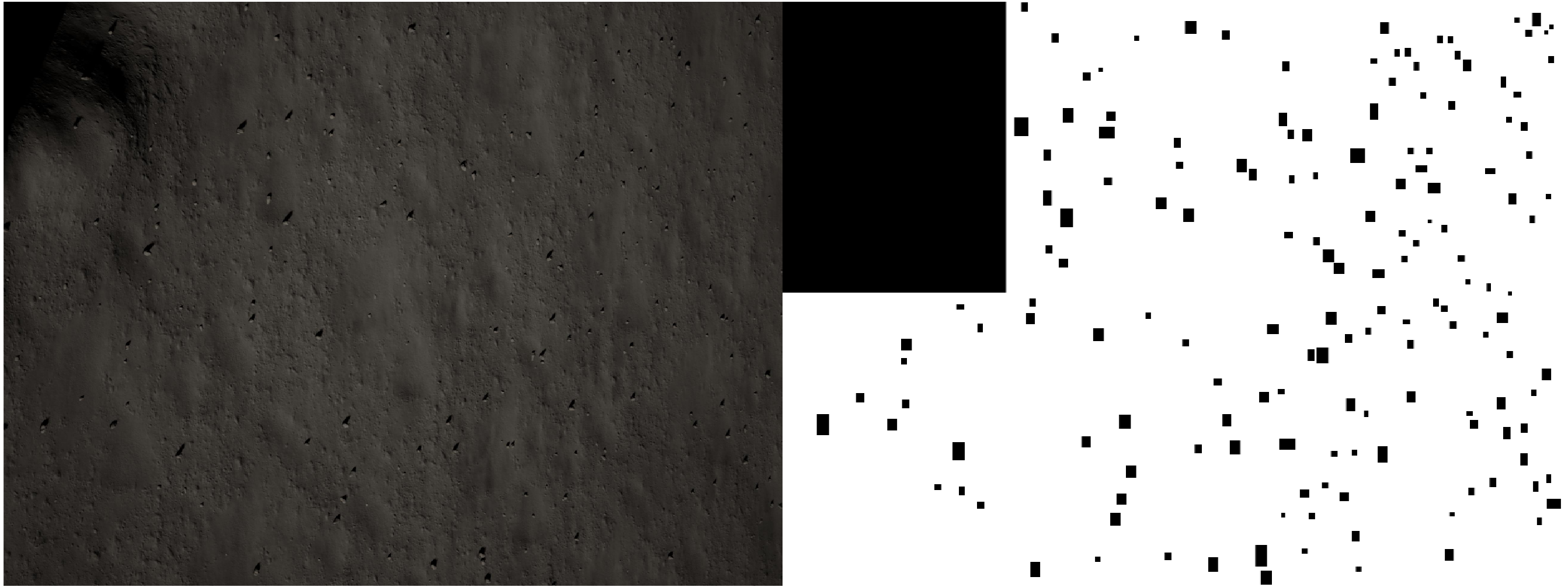}
	\caption{Example of hazard mask (rocks and craters) to filter safe landing sites.}
	\label{fig:mask:example}
\end{figure}

Once the Neural Network (NN) detects the hazards, a mask can be created to map areas that are good or bad. Good area selection can be augmented by subdividing the mask using a Quadtree algorithm to quickly identify areas away from the hazards. The approximate size of the pixels can be determined using the camera intrinsics to verify that area of those regions of interest is greater than or equal to the VFDE. Then, a classical structure from motion pipeline using the attitude and the camera parameters can be used to build a point cloud. This point cloud can be segmented using the hazard mask, to verify slope and roughness in the areas denominated good by fitting a plane using a simple least squares approach. An example of the hazard mask can be seen in Figure \ref{fig:mask:example}.

\begin{figure}[htb!]
    \centering
    \begin{minipage}[t]{0.4\linewidth}
        \centering
        \includegraphics[width=\linewidth,angle=90]{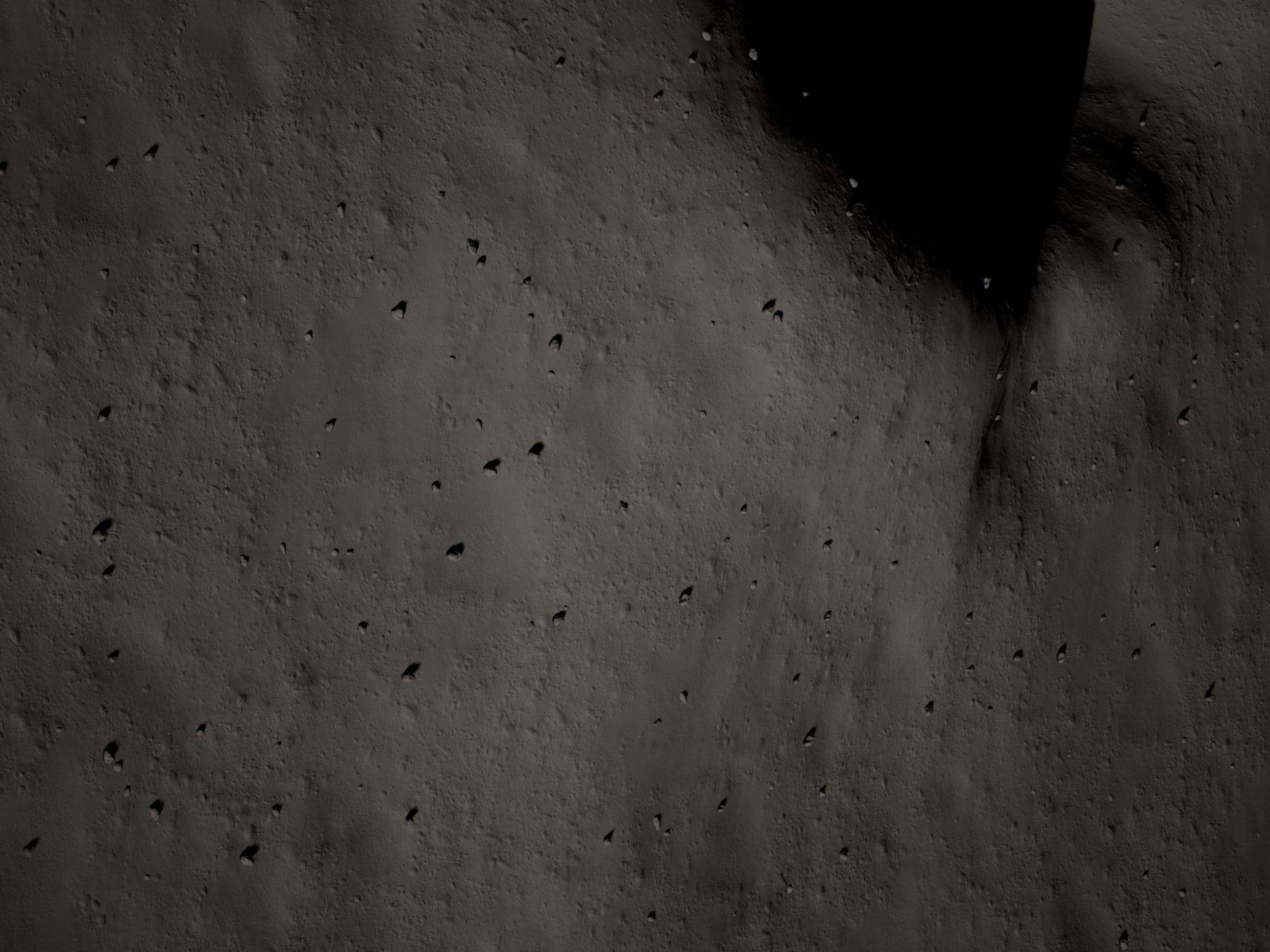}
    \end{minipage}
    \begin{minipage}[t]{0.59\textwidth}
        \centering
        \includegraphics[width=\linewidth]{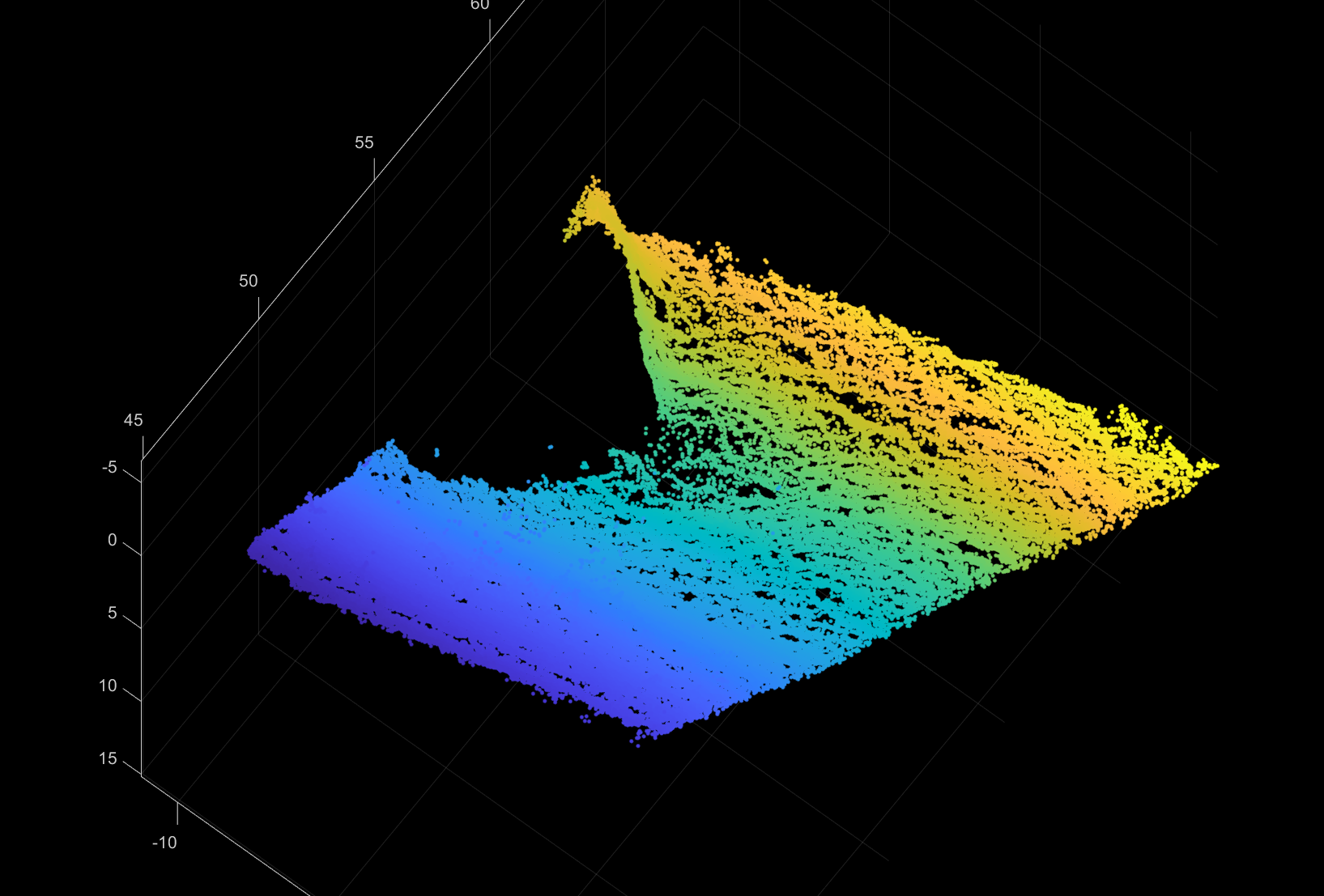}
    \end{minipage}
    \caption{Left: Lunar surface. Right: Point cloud reconstructed using structure from motion\cite{posada2020a}.}
    \label{fig:sfm:example}
\end{figure}

The sparse point cloud is generated by matching and triangulating features\cite{lindstrom2010triangulation}. These features are obtained by finding unique identifiers in the image using a feature detection algorithm. Multiple known methods such as SIFT, SURF, Fast, are used in robotics; however, for this approach ORB was used due to its feature detection, descriptor resilience, and computation speed\cite{karami2017image}. Once these features are identified in each region of interest, they can be triangulated using epipolar geometry mapping the pixel space into the 3-D space. Figure \ref{fig:sfm:example} shows an example of the point cloud reconstruction on right side and the lunar surface from the imagery acquired. The reconstruction is trapezoidal due to the camera geometry and it can be projected from the 3-D ILS frame into the camera frame and then into the camera frame using the projection matrix and the camera intrinsics. This is a well known problem in multiple view geometry and many examples can be found in the literature\cite{Hartley2004}.

Next step is to fit plane to measure the slope of the desired region. A simple least squares can be used to find the plane coefficients and therefore the vector that can provide the slope w.r.t the local gravity vector.\cite{posada2020a}
\begin{equation*}
    G\Vec{p} = \Vec{d}
\end{equation*}
\begin{equation*}
    \Vec{p} = (G^{T}G)^{-1}G^{T}\Vec{d}
\end{equation*}
\begin{equation}
    \left[\begin{matrix}a & b & c\end{matrix}\right]^{T} = (G^{T}G)^{-1}G^{T}\Vec{d}
\end{equation}
where $G$ is the point cloud matrix, and $d$ is a unitary column vector.

Once the coefficients are obtained, the angle $\beta$ between the local gravity vector ($g=[\begin{matrix}0&0&1\end{matrix}]^{T}$) and the normal vector $\Vec{p}$ is obtained using the dot product rule:
\begin{equation*}
    \beta = cos^{-1}\left(\frac{|\Vec{p}^{T}\cdot\Vec{g}|}{||\Vec{p}^{T}||\cdot||\Vec{g}||}\right)
\end{equation*}
\begin{equation}
    \beta = cos^{-1}\left(\frac{|c|}{\sqrt{a^2+b^2+c^2}}\right)
\end{equation}
If the the angle is larger than $90^\circ$ it can be corrected as:
\begin{equation*}
    \beta = 180^\circ - \beta
\end{equation*}
The least squares fitting error that represents the roughness of the surface $d_{e}$ can be calculated as:
\begin{equation}
    d_{e} = \frac{|ax_{k}+by_{k}+cz_{k}-1|}{\sqrt{a^2+b^2+c^2}},\quad k = 1,2, \dots, n
\end{equation}

\begin{figure}[h!]
    \begin{minipage}{0.5\textwidth}
        \centering
        \includegraphics[width=0.99\textwidth]{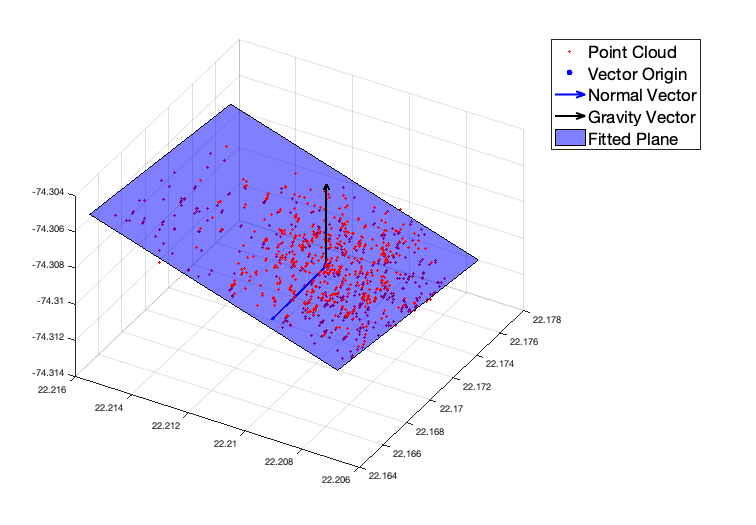}
    \end{minipage}%
    \begin{minipage}{0.5\textwidth}
        \centering
        \includegraphics[width=0.99\textwidth]{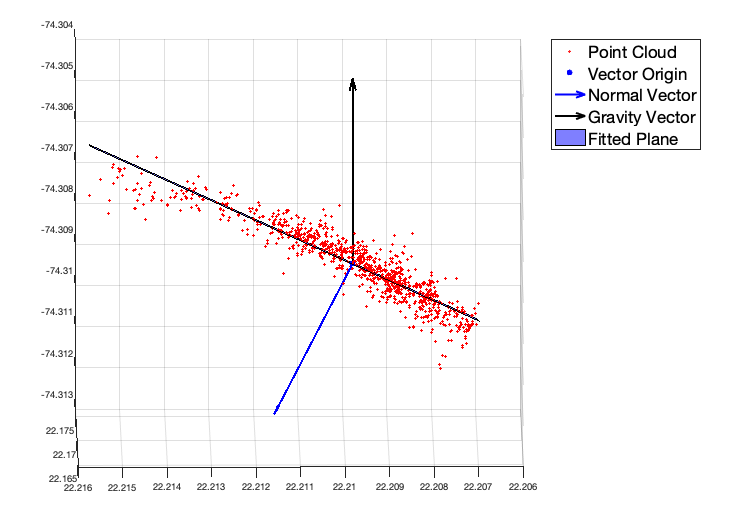}
    \end{minipage}
    \caption{Examples of plane fit on a region of interest.}
    \label{fig:dataset}
\end{figure}

Finally, a cost function can be minimized to rank the candidate sites based on the area, roughness, and slope. This can be augmented with the fuel ellipse depending on how much fuel the spacecraft will have when the algorithm is used. For a safety factor it is also assumed that the spacecraft is able to withstand a slope of $10^\circ$ before tilting and hazards up to 30 cm. The values in the cost function are normalized between 0 and 1, where closer to 0 is ideal and 3 is the worst.
\begin{equation}
\label{eq:cost}
    J = \frac{10m^2}{Area}+\frac{Slope}{10 ^\circ}+\frac{Roughness}{0.3 m}
\end{equation}

\section{Database}\label{S:Database}
\begin{figure}[htb!]
	\centering\includegraphics[width=\textwidth]{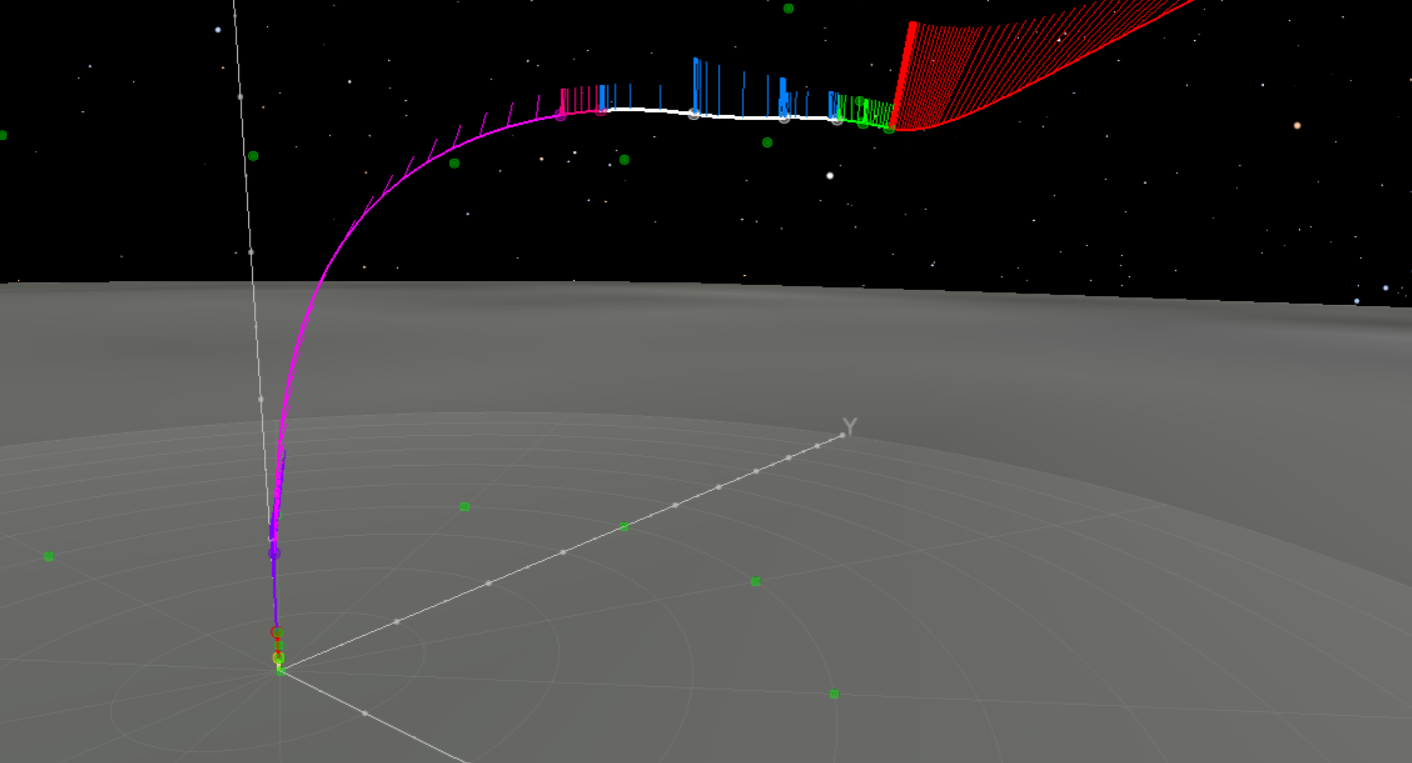}
	\caption{Ideal landing trajectory with respect to the intended landing site. The green highlight represents image collection for HDA. White represents the time HDA has to process and provide landing sites, as well as contingency for more image acquisition if needed.}
	\label{fig:hda_traj}
\end{figure}

To generate the imagery of the Database a trajectory for landing on the Moon was assumed with a specific path for performing site detection. Figure \ref{fig:hda_traj} illustrates an operational trajectory for HDA, which begins at 400 meters altitude and 400 meters downrange from the intended landing site (ILS). This trajectory allows approximately 5 seconds for image acquisition, and an additional 10 seconds for computation and relaying sites to the GNC system. Therefore the need for fast mobile implementation.

\subsection{Camera Parameters}
A pinhole camera is assumed for the purpose of this imagery as none of the images use a wide field of view (FOV) lens to limit the amount of radial and tangential distortion. The sensor is assumed to be a 5 MP camera with a FOV of $45^\circ$, resolution of 2592x1944 pix, and it is canted $45^\circ$ from the Nadir axis of the spacecraft looking directly at ILS in the lunar surface. The camera parameters are:
\begin{equation}
    K = \left[\begin{matrix}7363.60&0&1295.5\\0&7363.60&971.5\\0&0&1 
    \end{matrix}\right]
\end{equation}

This information can be used to estimate a scale that relates the meters per pixel to identify areas greater than or equal based on the pixel location on the image. 
\begin{equation}
\label{eqn:eqlabel}
\begin{split}
    & angLocation = \frac{pixLocation}{numPixels*FOV} 
    \\
    & \rho = \frac{altitude}{cos(camAngle-\frac{FOV}{2}+angLocation)}
    \\
    & res = \frac{2\sqrt{\rho^2-altitude^2}\,tan(\frac{FOV}{2})}{numPixels}
\end{split}
\end{equation}
where $altitude$ is the spacecraft z component w.r.t the ILS, $camAngle$ is the canted angle, $angLocation$ is the angular location of the pixel within the FOV, $pixLocation$ is the desired pixel location, $numPixels$ is the number of pixels along the image direction, $\rho$ is used to compute distance from Nadir to $pixLocation$, and $res$ is the final scaling resolution in meters per pixel to convert the side of the region of interest. For the intent it is assumed that the scale of the region of interest is the same and not going to vary much from the actual centroid.

\subsection{Synthetic Images}
Training and validation of the model depends on the database. Therefore, the synthetic imagery for testing HDA algorithms is built using high resolution synthetic digital elevation maps (DEMs), physically-based materials, artificial hazards, and dynamic lighting conditions\cite{getch2022}. A 40 x 40 km height map with 0.5 mpp (meters per pixel) resolution is imported into Unreal Engine 5 (UE5) and placed in the defined lunar-centered lunar-fixed (LCLF) location and orientation for rendering. 
Physically based materials are applied to the model using UE5's material blueprint shader graph to mimic a realistic lunar regolith. The lunar surface material is composed of an RGB base color and a roughness parameter that controls light reflectivity of the surface. These DEMs are enhanced for the analysis and hazards such as rocks and craters are added and modeled using known mathematical distribution\cite{li2017rock}. Some examples of this database can be seen in Figure \ref{fig:dataset}.

\begin{figure}[h!]
    \begin{minipage}{0.5\textwidth}
        \centering
        \includegraphics[width=0.99\textwidth]{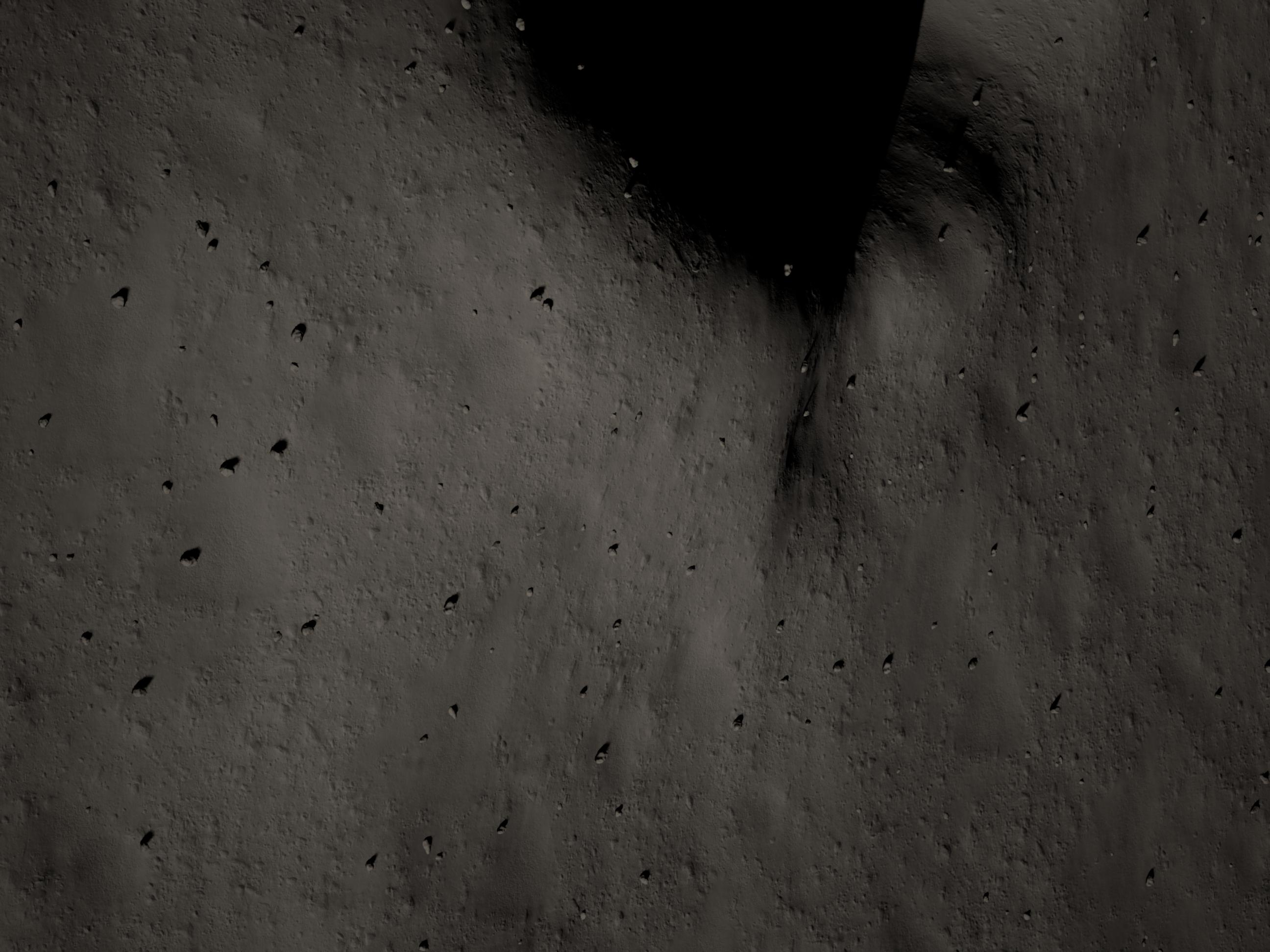}
    \end{minipage}%
    \begin{minipage}{0.5\textwidth}
        \centering
        \includegraphics[width=0.99\textwidth]{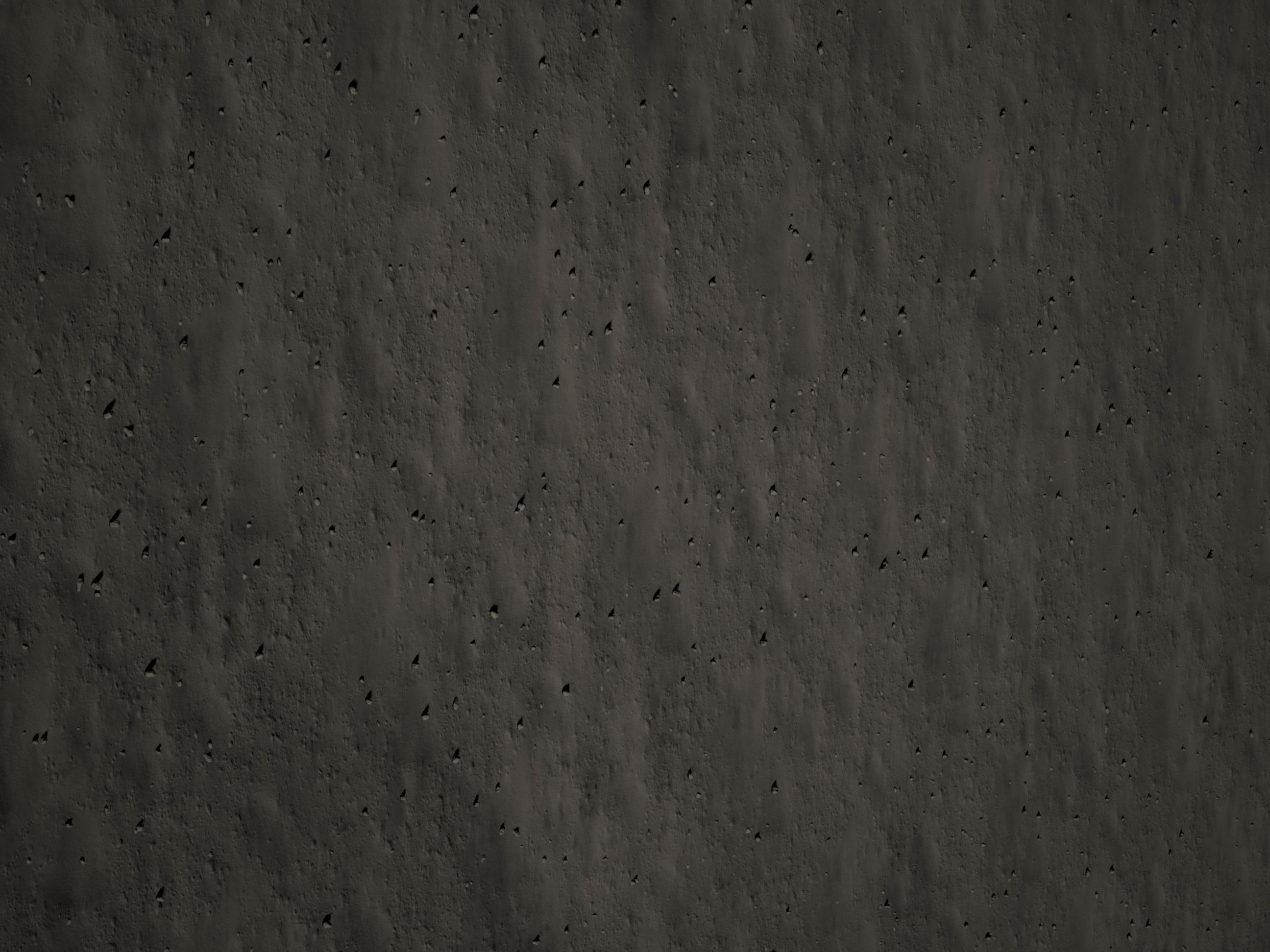}
    \end{minipage}
    \caption{Examples of dataset with different size hazards such as rocks and craters.}
    \label{fig:dataset}
\end{figure}

For assessing lunar landing site several classes were identified: rocks, shadows, craters, and safe spots. Images were picked random with different contrasts of a database of more than 300 images. Images were then labeled using free software labelImg using the selected classes ``rocks'' and ``crater''. where these features were outlined by bounding boxes. An example of a tagged image from the database is demonstrated in the Appendix in Figure \ref{fig:labeling}.

\section{Embedded Systems Neural Networks}
\begin{figure}[h!]
	\centering\includegraphics[width=\textwidth]{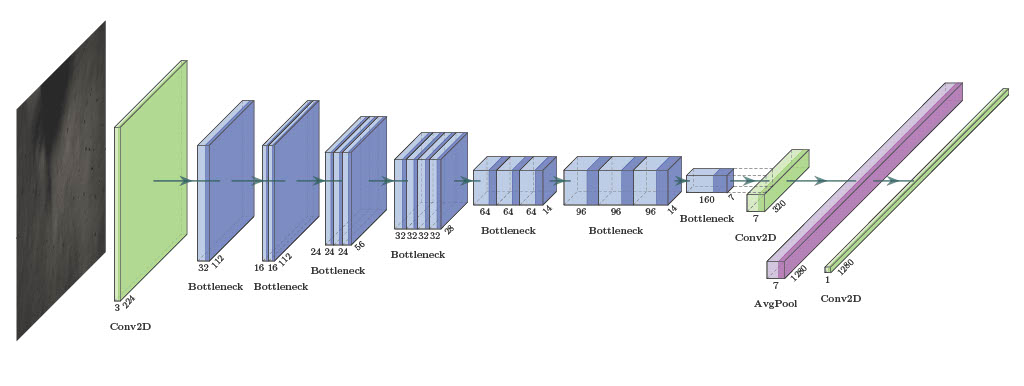}
	\vspace{-0.8cm}
	\caption{MobileNetV2 NN architecture before SSD and FPN.}
	\label{fig:mobiletnet}
\end{figure}

One of the main challenges of choosing or designing a NN for a space application is the size of the network, the number of hyper-parameters, portability, and the compute capability for a computationally-constrained mobile hardware. Currently available radiation hardened hardware is still many generations back having sometimes just one core and no parallel processing \cite{ginosar2012survey}. Some of the state-of-the-art mobile NN for single image detection are YOLOV5 ``You-Only-Look-Once'' (Recently V6 and V7 were released), Single Shot Detectors (SSD) based, and Feature Pyramid Network (FPN) based NNs\cite{sultana2020review}. Each NN has a trade-off between accuracy and latency. To achieve accuracy. portability, and speed MobileNetV2 NN was chosen as the core network (Figure \ref{fig:mobiletnet}) to decompose the image\cite{chiu2020mobilenet}. This architecture can be utilized for a variety of tasks, including classification, segmentation, and detection; however, the latter is the emphasis of this study. Then an SSD and FPN NN architecture is used to augment MobileNetV2 to identify the hazards and extract their features at different scales using a pyramid levels. This network has been tested before effectively in embedded systems with an effective inference time and a simple deployment. \cite{posada2022autonomous,jordan2022sat}

\section{Training and Validation}
\begin{figure}[h!]
    \begin{minipage}{0.5\textwidth}
        \centering
        \includegraphics[width=0.9\textwidth]{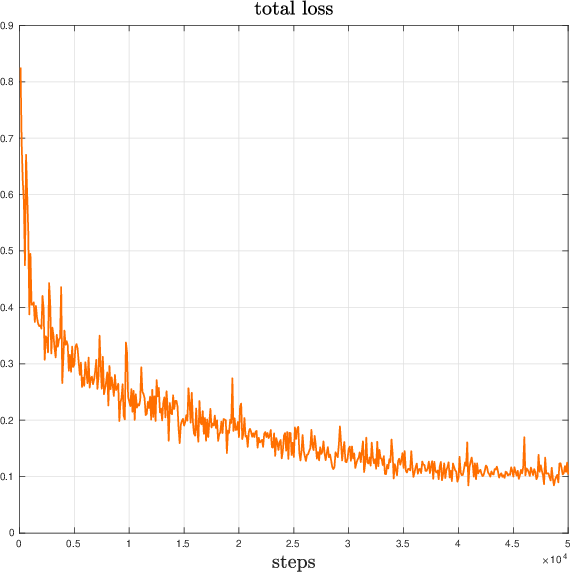}
    \end{minipage}%
    \begin{minipage}{0.5\textwidth}
        \centering
        \includegraphics[width=0.9\textwidth]{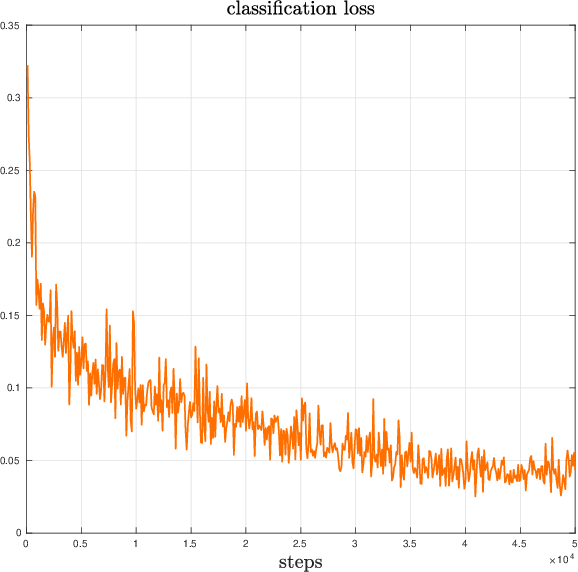}
    \end{minipage}
    \begin{minipage}{0.5\textwidth}
        \centering
        \includegraphics[width=0.9\textwidth]{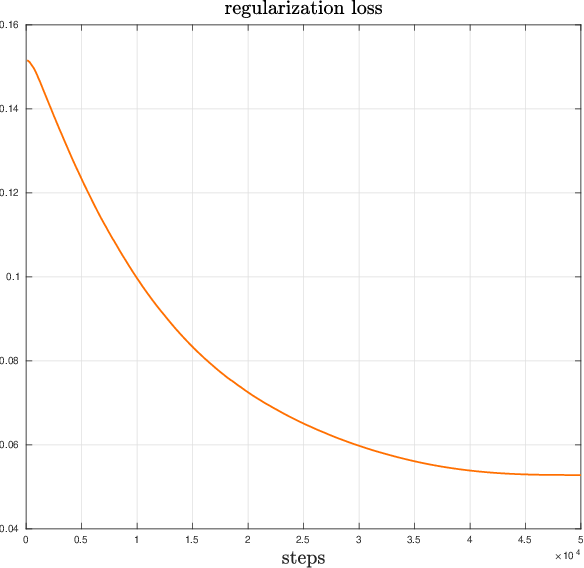}
    \end{minipage}%
    \begin{minipage}{0.5\textwidth}
        \centering
        \includegraphics[width=0.9\textwidth]{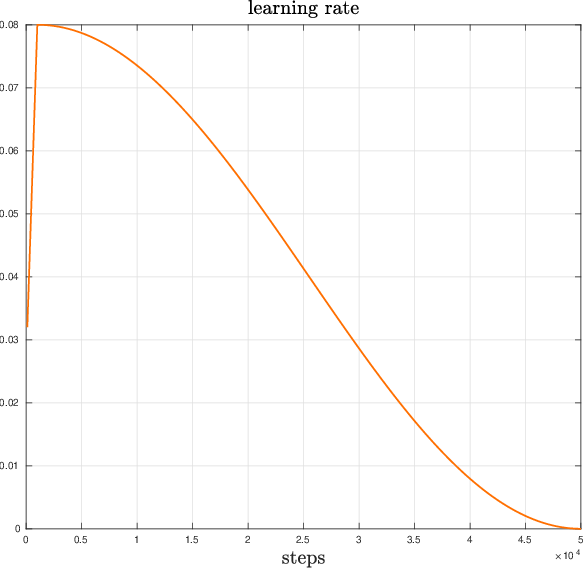}
    \end{minipage}
    \caption{Training output of cost functions. Total Loss, Classification Loss, Regularization Loss, and Learning Rate.}
    \label{fig:training:}
\end{figure}

MobileNetV2 network was training is done using the custom database described in the previous section. The NN training is tweaked to take the image at full-resolution and resize it to 1024 but retain the aspect ratio of the original image. Then a batch size of 2 was chosen in order to not rush the convergence of the learning and improve accuracy on detection over 50000 steps. The training was conducted using a TensorFlow-Python workflow on a workstation with two Intel Xeon Silver 4214 processors, an NVIDIA Quadro K6000, and 96 GB of system RAM. Figure \ref{fig:training:} shows the decrease in the different cost functions proving that the model prediction becomes better to detect the hazards. For validation, the network was evaluated in the computer it was trained as well on a Raspberry Pi 4B+ using the augmented dataset. The scripts are written in python using TensorFlow's version 2.4 and OpenCV's version 4.5.

\begin{table}[h!]
\caption{Approximate average time for detection inference for a 100 runs.}
\begin{center}
\resizebox{0.6\textwidth}{!}{%
\begin{tabular}{lclll}
\multicolumn{1}{c}{\begin{tabular}[c]{@{}c@{}}Validation\\ Hardware\end{tabular}} & \begin{tabular}[c]{@{}c@{}}Average Inference \\ Time ($sec$)\end{tabular} &  &  &  \\ \cline{1-2}
\begin{tabular}[c]{@{}l@{}}Desktop PC\\ Intel Xeon + Quadro K6000\end{tabular} & 3.223 &  &  &  \\
Raspberry Pi 4B+ & 2.867 &  &  &  \\
 & \multicolumn{1}{l}{} &  &  & 
\end{tabular}   
}
\end{center}
\label{tab:inference:time}
\end{table}
\section{Results}
\begin{figure}[h!]
    \begin{minipage}{0.5\textwidth}
        \centering
        \includegraphics[width=0.99\textwidth]{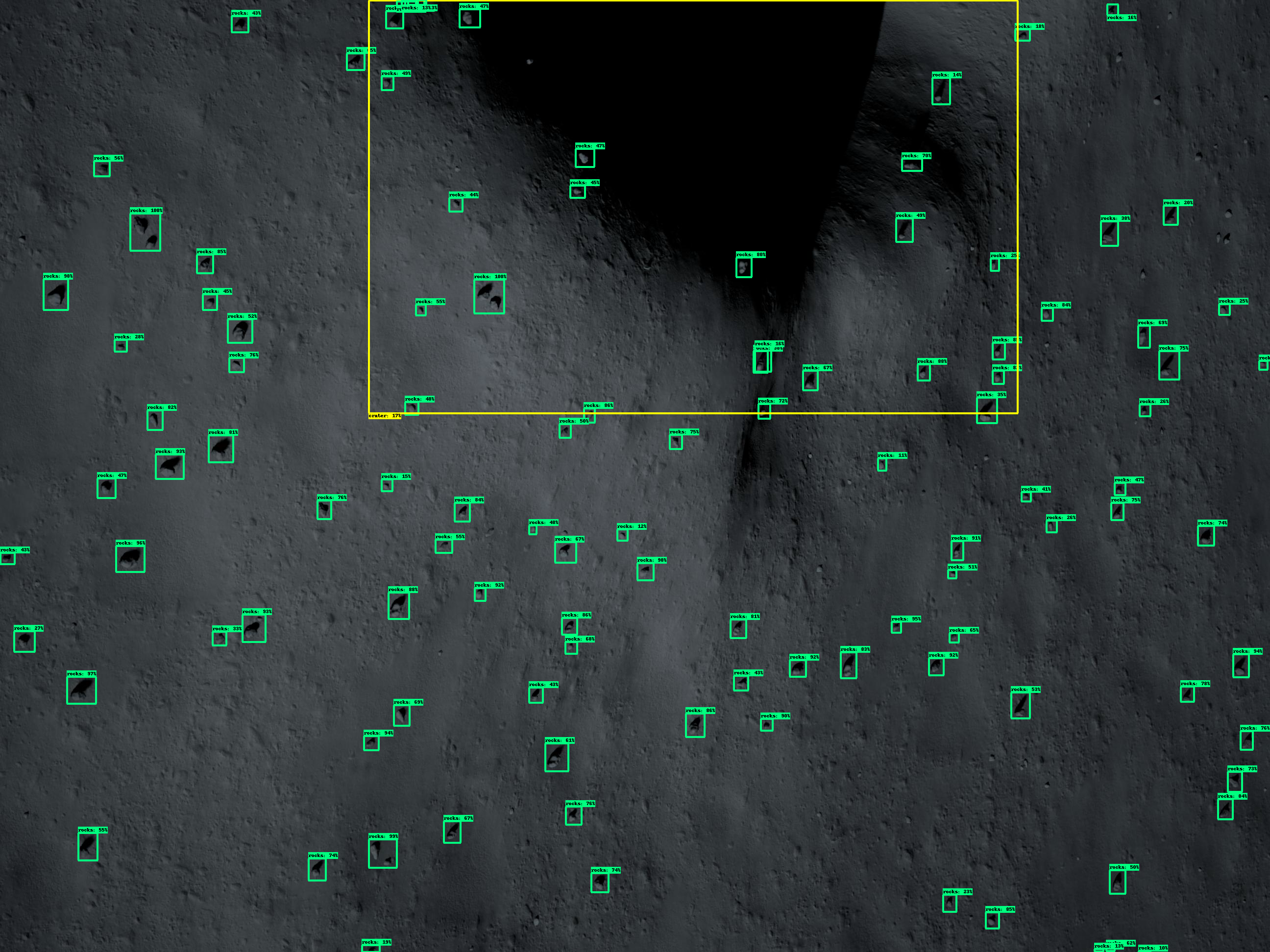}
    \end{minipage}%
    \begin{minipage}{0.5\textwidth}
        \centering
        \includegraphics[width=0.99\textwidth]{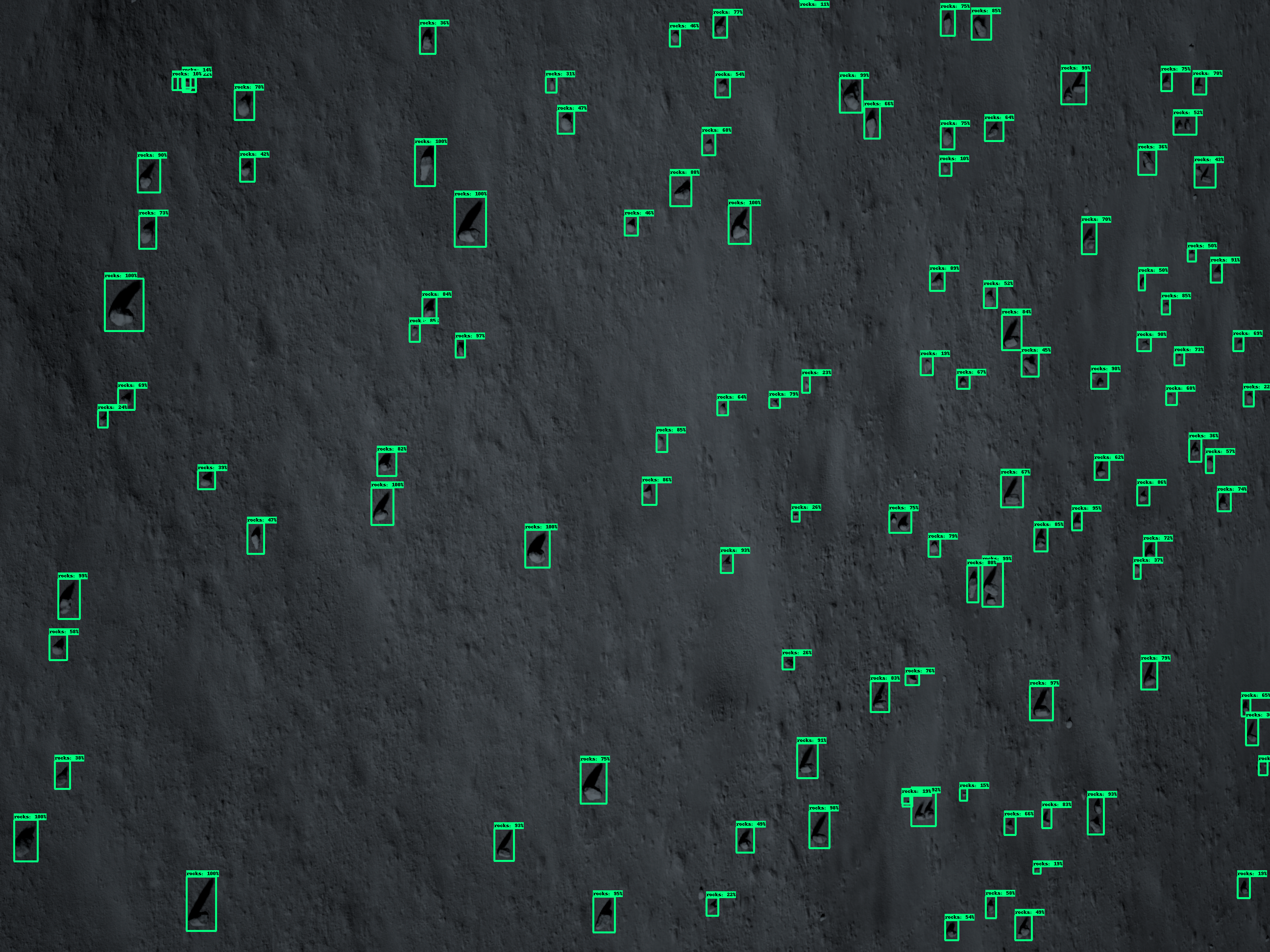}
    \end{minipage}
    \caption{Output of the NN for different imagery running inference on the desktop computer. Green boxes are detections of rocks and shadows, Yellow boxes are detections of craters.}
    \label{fig:results}
\end{figure}

\begin{figure}[h!]
    \begin{minipage}{0.5\textwidth}
        \centering
        \includegraphics[width=0.99\textwidth]{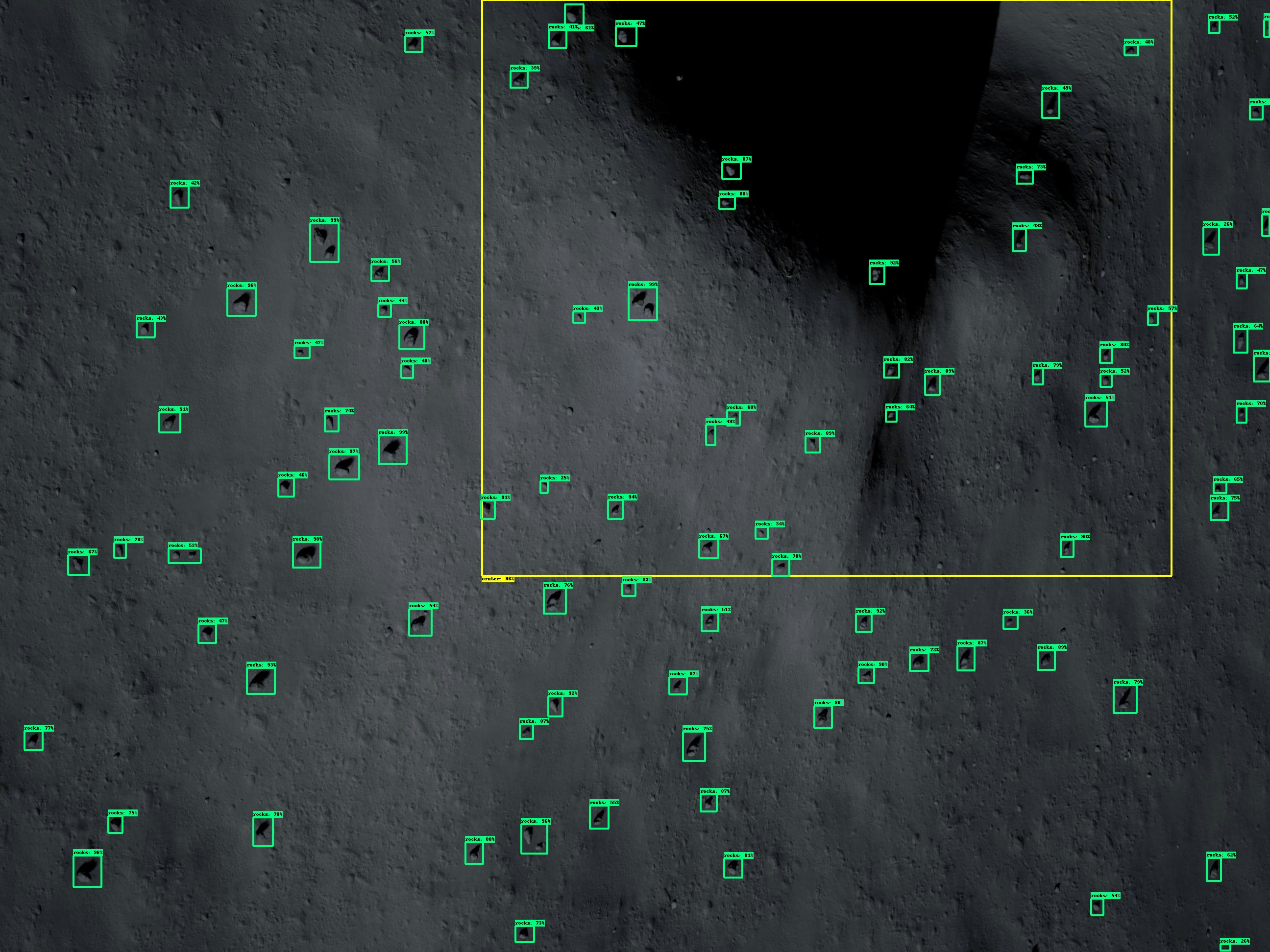}
    \end{minipage}%
    \begin{minipage}{0.5\textwidth}
        \centering
        \includegraphics[width=0.99\textwidth]{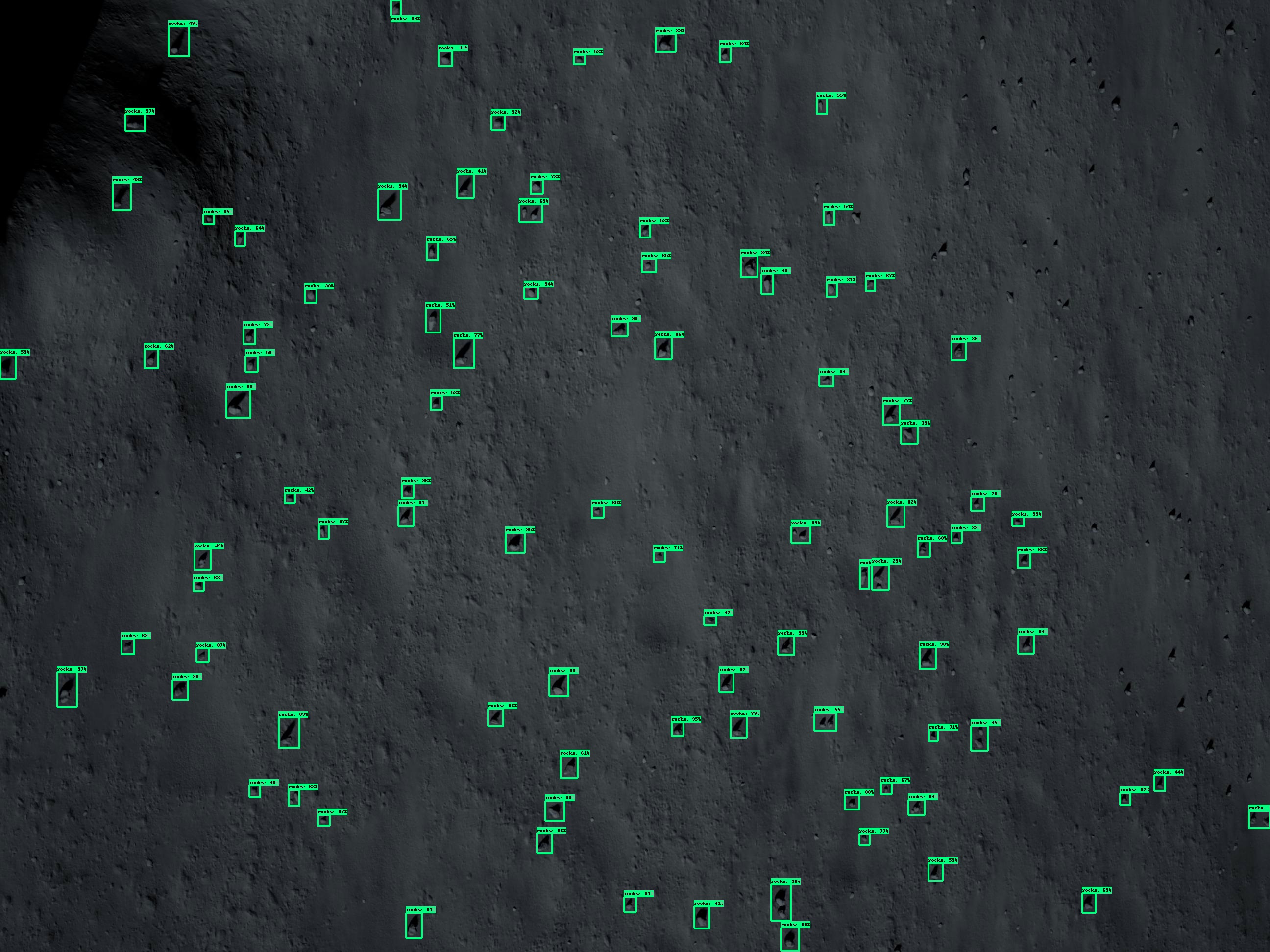}
    \end{minipage}
    \caption{Output of the NN for different imagery running inference on the Raspberry Pi. Green boxes show positive detection of rocks and shadows, Yellow boxes identify craters.}
    \label{fig:results:pi}
\end{figure}

\begin{figure}[h!]
	\centering\includegraphics[width=0.9\textwidth]{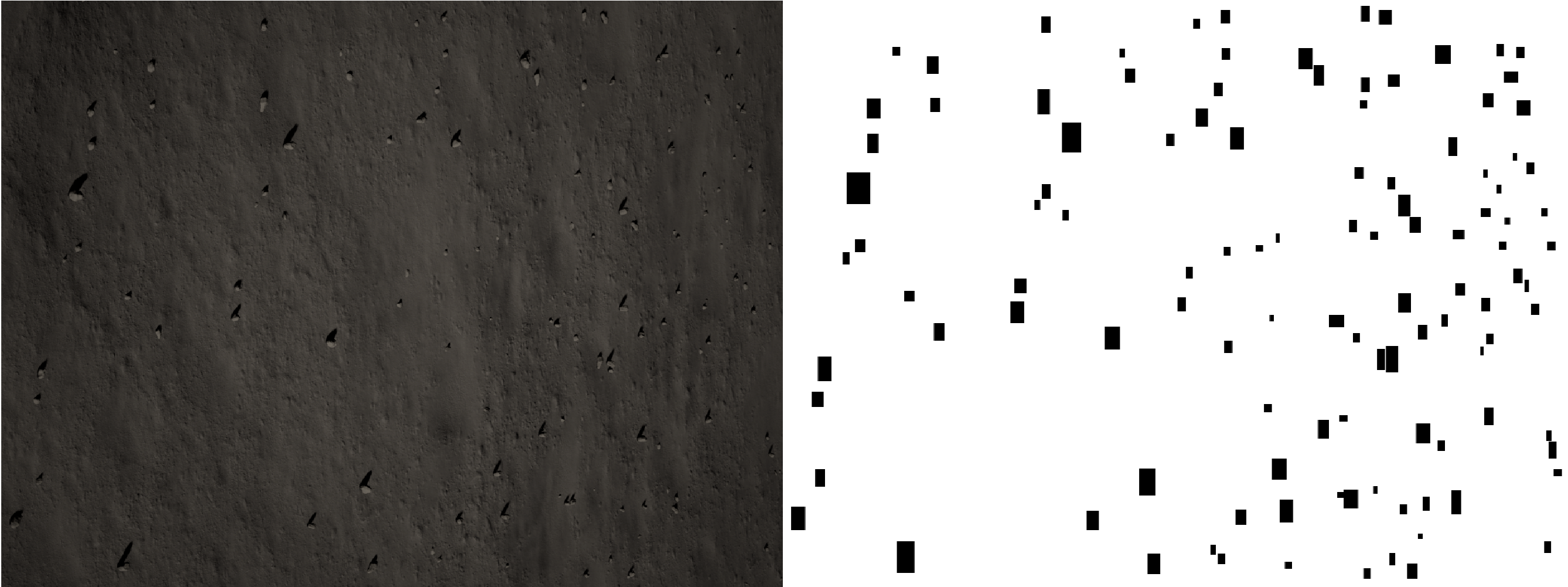}
	\caption{Hazard mask from right top corner in Figure \ref{fig:results}}
	\label{fig:mask:real}
\end{figure}

\begin{figure}[h!]
	\centering\includegraphics[width=0.45\textwidth]{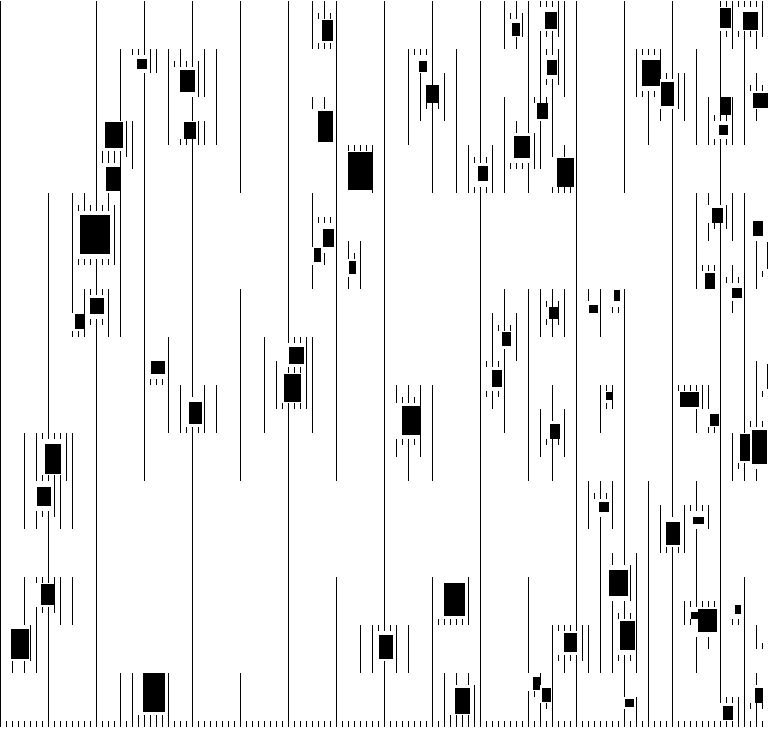}
	\caption{Quadtree decomposition to find sites relative to the hazards. Area is then verified to make sure the VFDE matches.}
	\label{fig:Quadtree}
\end{figure}

After training Figure \ref{fig:results} and Figure \ref{fig:results:pi} display the results of the NN running inference on the desktop computer and the Raspberry Pi 4B+ respectively.
In most of the detection cases the network was able to identify almost every hazard, some small rocks can be passed as noise depending on the texture of the surface. However, this could be addressed by increasing the dataset with more contrast enhancements and more brightness conditions. A stress case with more than 250 hazards was tested on the Raspberry Pi to test inference time. The right image from Figure \ref{fig:results:pi} illustrates this stress case where in the far FOV the rocks are not detected, but inference happened in $\approx2.9$ secs. 
Once the detections are obtained, a binary mask is generated as seen in Figure \ref{fig:mask:real}, which is then augmented using the Quadtree algorithm (Figure \ref{fig:Quadtree}). This mask and augmentation runs in $\approx 100$ msecs as the mask is binary and easier to process. When the mapping is done, ideal candidates based on the area are selected. To obtain the approximate size the of the area, Equation (\ref{eqn:eqlabel}) is used and yields an approximate value of 128 pix for a 10 m side VFDE.

\begin{figure}[h!]
	\centering\includegraphics[width=0.4\textwidth]{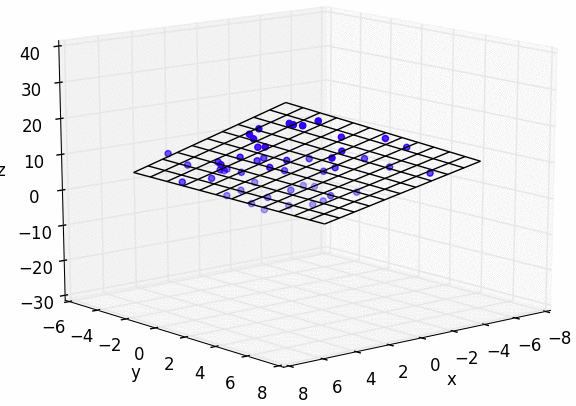}
	\caption{Plane fit of point cloud reconstructed from the top left corner region of interest.}
	\label{fig:plane:sample}
\end{figure}

These areas are used to delimit the the point cloud to analyze for slope locally on each region of interest. Figure \ref{fig:plane:sample} illustrates a plane fitted to the point cloud from the top left corner with a slope of $\approx 2.3$ deg.

\begin{figure}[h!]
	\centering\includegraphics[width=\textwidth]{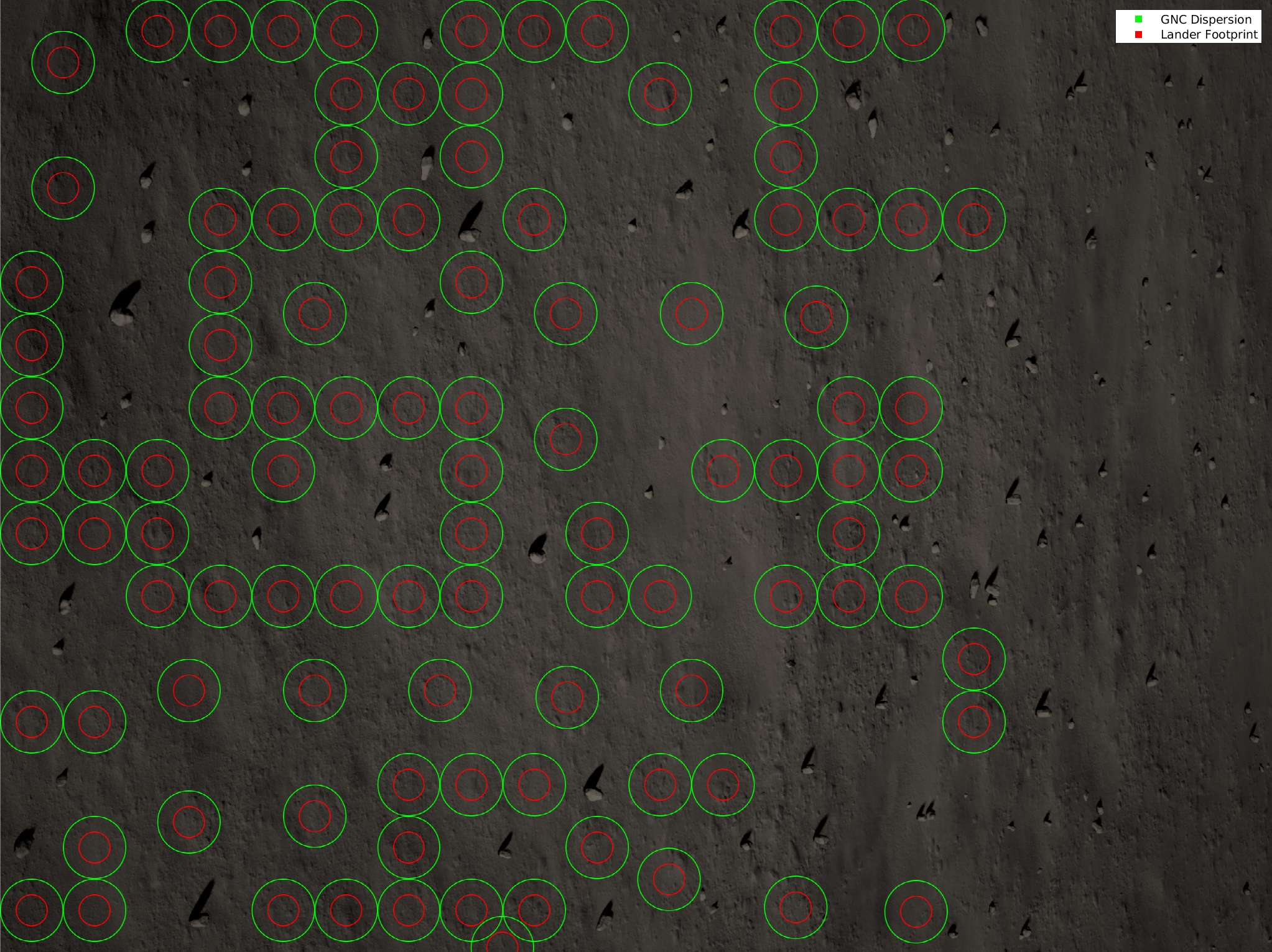}
	\caption{Green is a VFDE with the GNC dispersion and red is the spacecraft size.}
	\label{fig:SafeSites}
\end{figure}

Figure \ref{fig:SafeSites} illustrates all the safe sites in the location in close and mid FOV. The green circles are the VFDE that can be fitted into the desired regions; however, the real area identified in the Quadtree is the one used in a cost function that minimizes and ranks the sites based on area, roughness, and slope. The larger and cleaner the area, the safer it is for the spacecraft to land.

\section{Conclusions and future work}
Using NN for detecting hazards is useful in constrained embedded environments. MobileNetV2 particularly is a good architecture for this type of devices as it can process almost full-resolution imagery. A Raspberry Pi which is a low-cost, low-power embedded device is able to generate a complete solution for safe landing in the lunar surface. The network training can be improved with an increased database spanning a set that is able to generalize landing conditions not only on the lunar environment, but also in asteroids, and other planets such as Mars or Venus. The current implementation relies on python and could be faster if properly optimized in C++. This system can be easily deployed and used with a simple of-the-shelf-component to augment the safety of a spacecraft when landing. The authors encourage the space community to be more open to test and use mobile embedded systems for this type of applications. Future work includes the generation of a larger dataset, optimization of code in lower level programming language such as C++, use GPUs to decrease processing times, and testing of more embedded hardware such as NVIDIA portfolio of mobile platforms.

\section*{Appendix: Additional Figures}
\begin{figure}[h!]
	\centering\includegraphics[width=\textwidth]{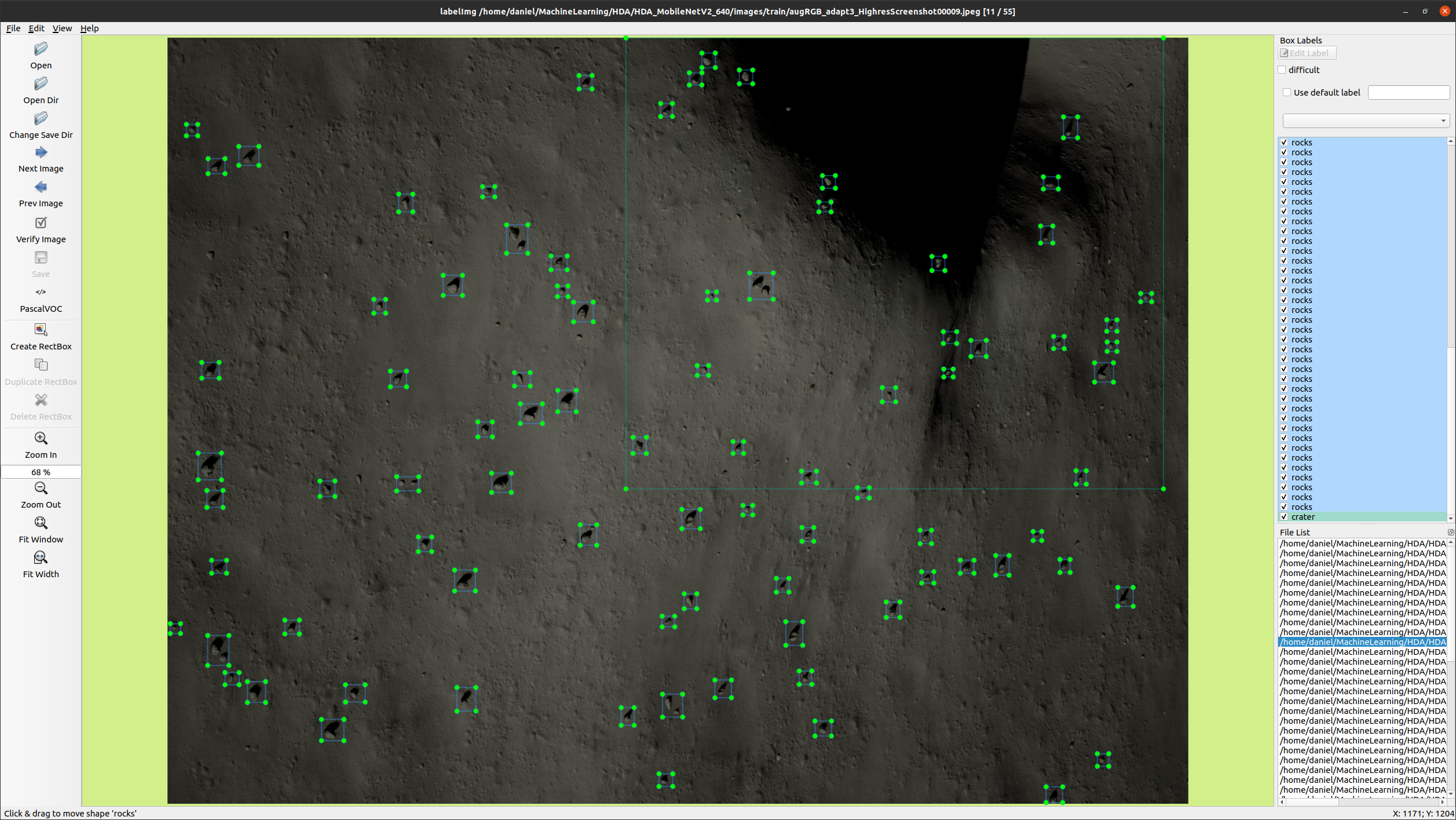}
	\caption{labelImg is used for labeling the database.}
	\label{fig:labeling}
\end{figure}

\begin{figure}[h!]
    \begin{minipage}{0.5\textwidth}
        \centering
        \includegraphics[width=0.99\textwidth]{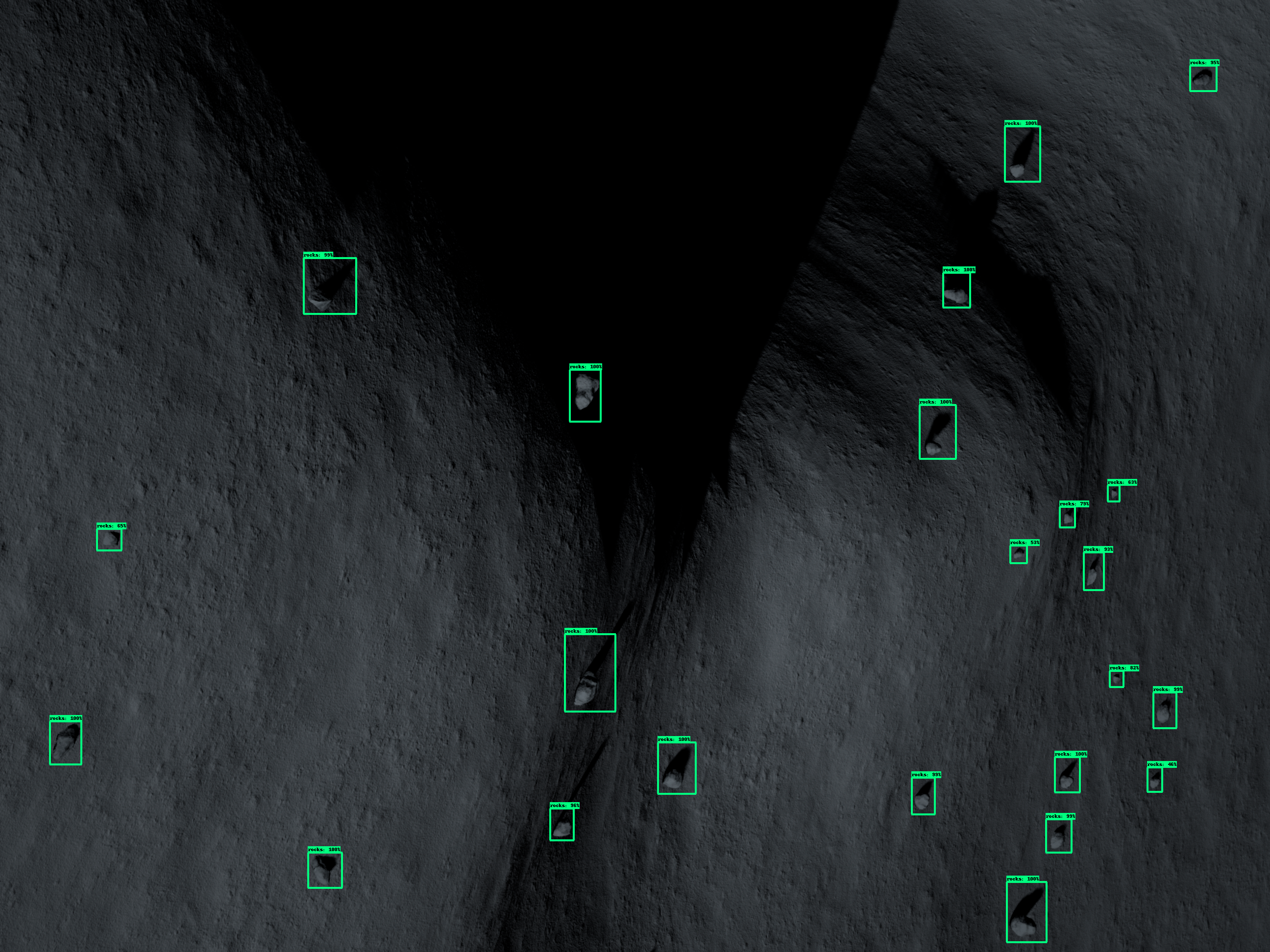}
    \end{minipage}%
    \begin{minipage}{0.5\textwidth}
        \centering
        \includegraphics[width=0.99\textwidth]{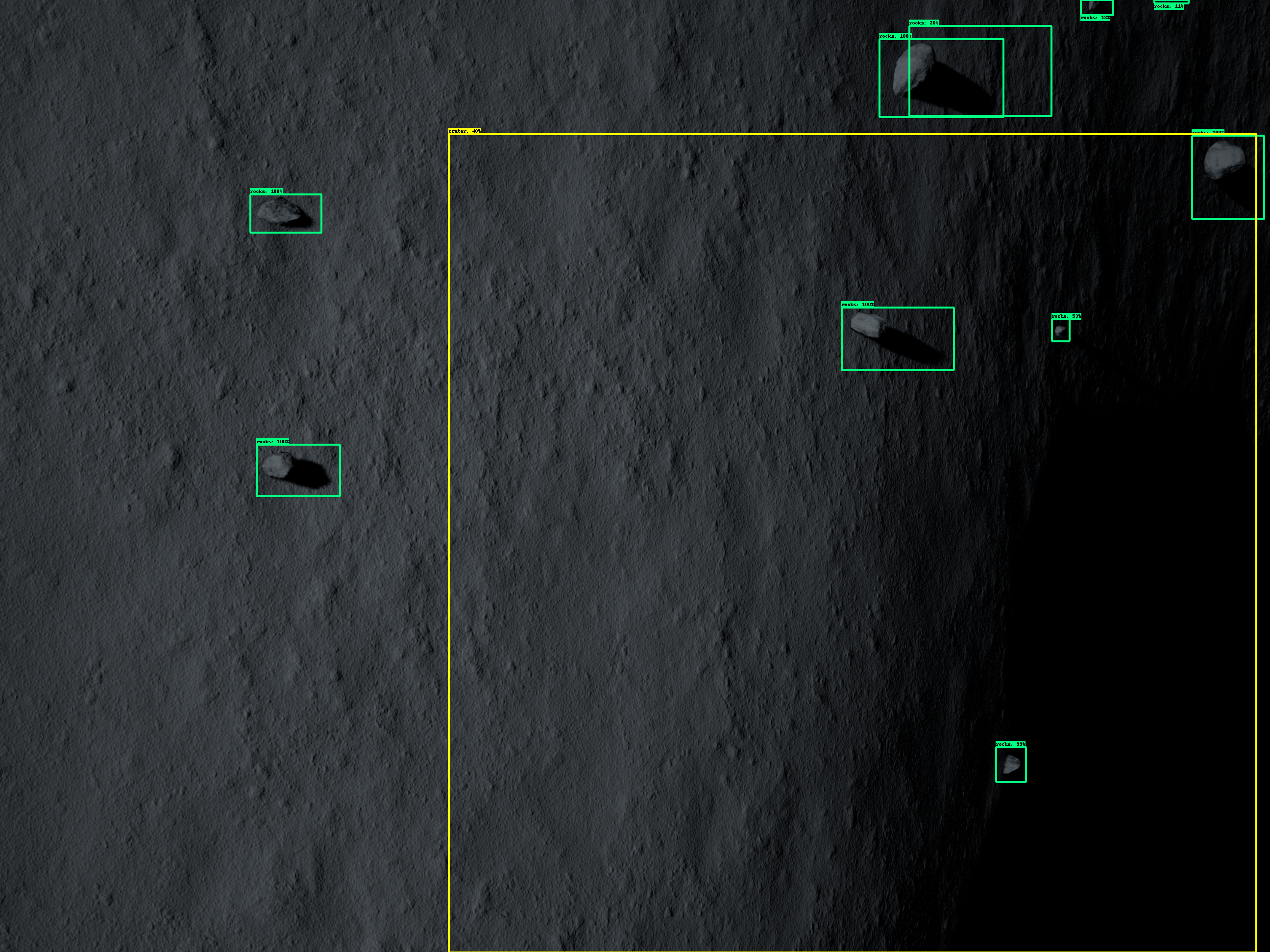}
    \end{minipage}
    \caption{Additional output of the NN for different imagery running inference on the desktop computer. Green boxes are detections of rocks and shadows, Yellow boxes are detections of craters.}
\end{figure}

\clearpage
\bibliographystyle{AAS_publication}   
\bibliography{hda_ML}   

\end{document}